\documentclass[10pt,journal]{IEEEtran}
\usepackage{graphicx}
\usepackage{amsmath}
\usepackage{booktabs}
\usepackage{caption}
\usepackage{subcaption}
\usepackage{hyperref}
\usepackage{url}
\usepackage{multirow}
\usepackage{float}
\usepackage{cite}

\begin{document}

\title{EGD-YOLO: A Lightweight Multimodal Framework for Robust Drone-Bird Discrimination via Ghost-Enhanced YOLOv8n and EMA Attention under Adverse Conditions}

\author{Sudipto~Sarkar, ~Mohammad~Asif~Hasan, ~Khondakar~Ashik~Shahriar, \\ Fablia~Labiba, ~Nahian~Tasnim, ~Shaikh~Anowarul~Haq~Fattah \\
Bangladesh University of Engineering and Technology, Dhaka-1000 \\
Email: sudiptosarkarjoy@gmail.com}

\maketitle

\begin{abstract}
Accurate and effective identification of drones versus birds is essential for safeguarding airspace integrity, advancing surveillance operations, and strengthening security measures in fast-evolving aerial scenarios. Leveraging the VIP CUP 2025 dataset---with its synchronized RGB and infrared (IR) images---this study provides an ideal testbed for exploring advanced multimodal fusion approaches. Herein, we unveil EGD-YOLOv8n, an efficient yet potent object detection framework built upon YOLOv8n, which smartly evolves the C2f components into C3Ghost structures infused with GhostBottlenecks to deliver exceptional feature capture, enhanced multi-scale flexibility, and deeper semantic and contextual insights, ultimately elevating detection accuracy and reliability while significantly shrinking model footprint and operational expenses. Complementary optimizations include swapping out standard convolutions in the neck for GhostConvs to amplify efficiency, alongside embedding an Efficient Multi-scale Attention (EMA) layer ahead of C3Ghost units to promote inter-spatial knowledge transfer and sharpened channel- and spatial-feature refinement. Addressing the constraints of rigid, predefined kernel sizes in conventional architectures that falter against diverse object geometries, our redesigned detection head employs DDetect---a deformable convolution paradigm that leverages learnable offsets for dynamic shape adaptation, substantially fortifying detection resilience to morphological discrepancies. Three specialized detectors are developed and benchmarked: one for RGB inputs alone, one for IR alone, and a hybrid RGB-IR fusion variant. Rigorous testing on the VIP CUP 2025 validation set underscores the fused EGD-YOLOv8n's marked advancements in precision and mean average precision (mAP) over single-modality counterparts and existing benchmarks, without compromising sub-second inference latencies on everyday GPUs---heralding practical, adaptable tools for robust aerial hazard countermeasures.
\end{abstract}


\section{Introduction}
\label{sec:introduction}

The rise of unmanned aerial vehicles (UAVs), or drones, poses acute risks to airspace security, from unauthorized intrusions to collisions with manned aircraft and infrastructure disruptions. A pivotal yet challenging task is distinguishing drones from birds in real-time surveillance, as misclassification risks false alarms against avian flocks or undetected threats in restricted zones. Both exhibit small footprints, erratic paths, and camouflage in cluttered skies, compounded by imaging artifacts like speckle noise, motion blur, and camera jitter.

RGB-based systems capture structural details (e.g., propellers) but falter in low light, fog, or occlusions. IR thermography excels in adverse conditions via heat signatures yet struggles with overlapping thermal profiles for drone-bird differentiation. Multimodal RGB-IR fusion thus promises resilient detection by combining complementary strengths.

This work addresses these issues in the IEEE SPS Video and Image Processing (VIP) Cup 2025 at ICIP 2025, focusing on IR-visual fusion for drone detection and discrimination in distorted videos. The dataset features 30 videos (320×256 resolution, 30 fps) with paired RGB-IR frames across terrains (hilly, forested) and weathers (cloudy, foggy), applying distortions on a 1-5 intensity scale: speckle/salt-and-pepper noise (2), AWGN (3), Gaussian blur/uneven illumination/motion blur (4), and camera instability (5). Scenarios include isolated/multi-class targets, peripheral low-res objects, swarms, occlusions, and low light, evaluating metrics like mAP, F1-score, and robustness (R\_b) for detection continuity (<15 missed frames).

For the drone-bird detection subtasks, we propose EGD-YOLOv8n---a lightweight YOLOv8n variant exploiting RGB-IR fusion. It upgrades C2f to C3Ghost with GhostBottlenecks for enhanced feature extraction and multi-scale adaptability; replaces neck convolutions with GhostConvs, prefixed by EMA for cross-spatial refinement; and equips the head with DDetect deformable convolutions for adaptive offsets against shape variations.

Contributions include:

\begin{itemize}
\item[(i)] baselines for RGB-only, IR-only, and fused modalities assessing fusion gains;
\item[(ii)] a Ghost-EMA-Deformable framework for real-time ($\geq$30 fps) distortion resilience; and
\item[(iii)] evaluations on 45,000 training/6,500 validation images showing superior mAP and inference over baselines.
\end{itemize}

This advances multimodal UAV monitoring for secure, drone-dense airspace.

\section{Related Works}
\label{sec:related_works}

Drone detection has evolved rapidly with deep learning paradigms, particularly YOLO variants, offering real-time capabilities for aerial surveillance. Early works like YOLOv4 and YOLOv5 demonstrated efficacy in RGB-based drone localization, achieving high mAP on visible spectra but struggling with small, fast-moving targets under distortions. Subsequent advancements, such as YOLOv11 in dual-stream setups, parallelize RGB and IR processing for payload detection, attaining 45 FPS on edge devices while fusing thermal cues for nocturnal robustness. However, these often overlook bird discrimination, where morphological similarities yield false positives exceeding 20\% in cluttered scenes.

Distinguishing drones from birds remains a core challenge, addressed in grand challenges like the ICASSP 2023 Drone-vs-Bird Detection event, which reviewed 15+ methods emphasizing micro-Doppler signatures and bio-inspired motion cues for trajectory-based classification. Reviews highlight deep learning's superiority over classical approaches, with CNNs like ResNet achieving 92\% accuracy on segmented RGB datasets, yet faltering in low-light or occluded swarms. Bio-mimetic models, inspired by avian vision, enhance edge detection for erratic flights, reducing misclassifications by 15\% in multi-class frames. Out-of-distribution strategies, including energy-based models, further bolster generalization against environmental variances.

Multimodal RGB-IR fusion mitigates unimodal limitations, leveraging RGB's textural fidelity with IR's thermal resilience. Pioneering efforts generated synthetic IR via CycleGAN for YOLO training, yielding 85\% mAP on fused pairs. Recent innovations, like GLFDet's global-local optimization, integrate attention mechanisms for cross-modal alignment, improving small-drone recall by 12\% in hilly terrains. Datasets such as multi-perspective RGB-IR collections simulate VIP Cup distortions, validating fusion gains in fog and motion blur. Lightweight networks like PONet fuse at the feature level for vehicle analogs, adaptable to UAVs with <10M parameters.

Despite these strides, gaps persist in distortion-resilient, lightweight architectures for drone-bird tasks. Prior YOLO fusions prioritize accuracy over efficiency, rarely incorporating deformable convolutions or Ghost modules for parametric compression. Our EGD-YOLOv8n bridges this by embedding C3Ghost, EMA, and DDetect in a multimodal pipeline, tailored to VIP Cup 2025's severe degradations (e.g., intensity-5 camera instability), surpassing baselines in mAP while sustaining $\geq$30 FPS.

\section{Methodology}
\label{sec:methodology}

\subsection{Dataset Description}

The drone-bird detection task in the IEEE SPS VIP Cup 2025 is based on a large-scale multimodal dataset consisting of paired RGB and infrared (IR) images captured under diverse surveillance scenarios. The dataset contains a primary annotation category focused on drone versus bird classification, with approximately 45,000 training images, 6,500 validation images, and 13,000 test images. These annotations emphasize binary discrimination between drones and birds, with balanced class distributions to support robust learning: the RGB subset includes around 30,000 bird instances and 36,000 drone instances, while the IR subset features 31,000 bird instances and 37,000 drone instances. This near-equilibrium setup minimizes bias and enables models to capture subtle discriminative cues, such as structural rigidity in drones versus organic motion in birds.

\begin{figure}[H]
\centering
\includegraphics[width=0.45\textwidth]{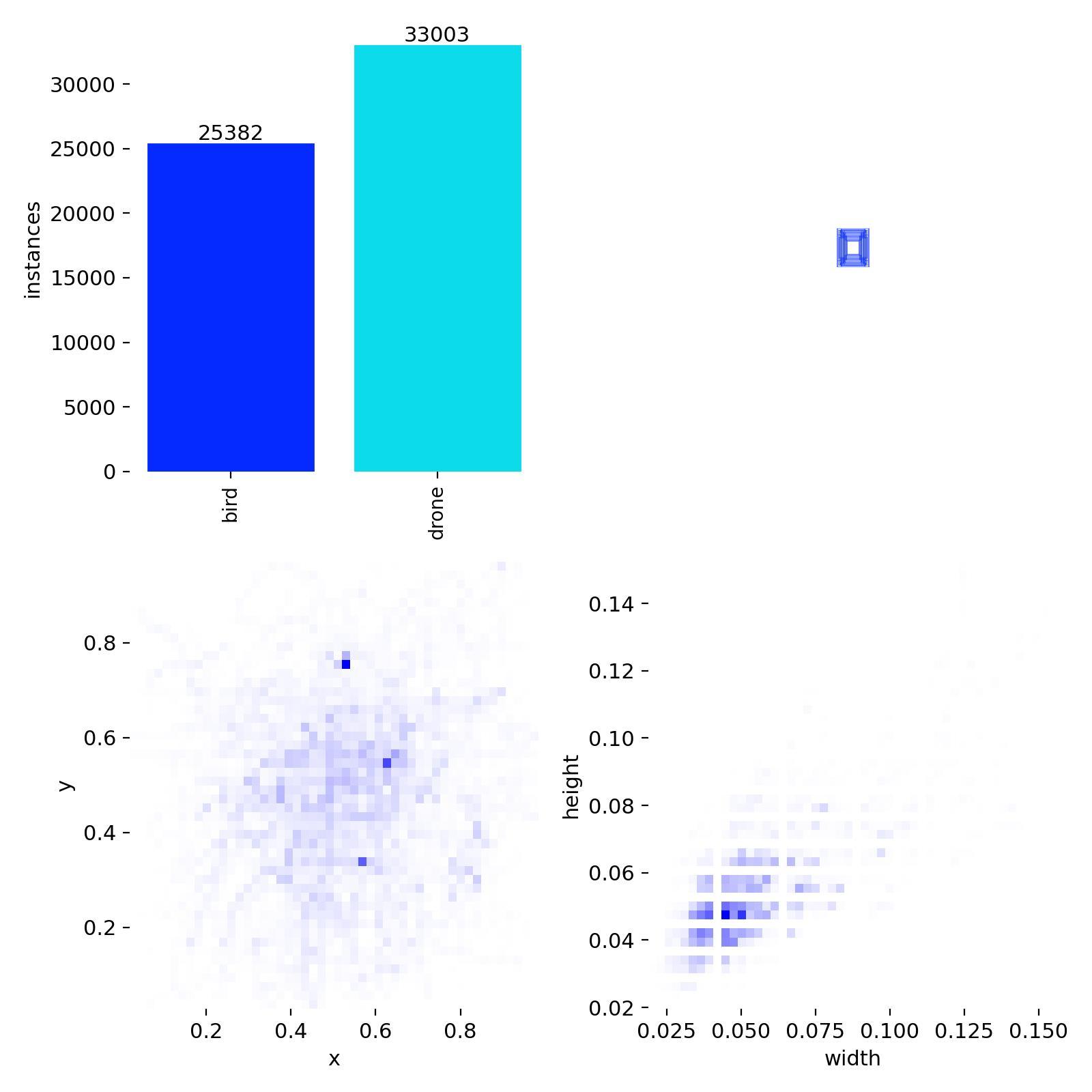}
\caption{Detection Data Distribution}
\label{fig:data_distribution}
\end{figure}

The slight overrepresentation of drone labels in both modalities reflects real-world priorities in airspace monitoring, where UAV threats warrant enhanced focus. All images are captured at a native resolution of 320 × 256 pixels and incorporate a wide array of environmental and distortion factors to mimic operational challenges in aerial surveillance. Environmental variations encompass urban clutter, rural landscapes, low-altitude fog, high-altitude mist, and variable lighting from dawn to dusk. Distortions simulate real-time capture issues, including Gaussian noise, motion blur from camera shake or object velocity, salt-and-pepper noise, uneven illumination gradients, and additive white Gaussian noise (AWGN), applied at intensity levels up to 5/5 for severe cases like rapid aerial maneuvers or atmospheric turbulence. These elements collectively test model generalization, forcing reliance on multimodal cues to resolve ambiguities like thermal overlaps between bird flocks and drone swarms.

Overall, the dataset serves as a rigorous benchmark for multimodal drone surveillance, facilitating advancements in detection accuracy amid dynamic, low-signal environments. The paired RGB-IR structure uniquely supports fusion techniques, allowing exploitation of RGB's textural fidelity alongside IR's illumination-invariant thermal mapping for superior bird-drone separation.

\begin{figure}[H]
\centering
\includegraphics[width=0.45\textwidth]{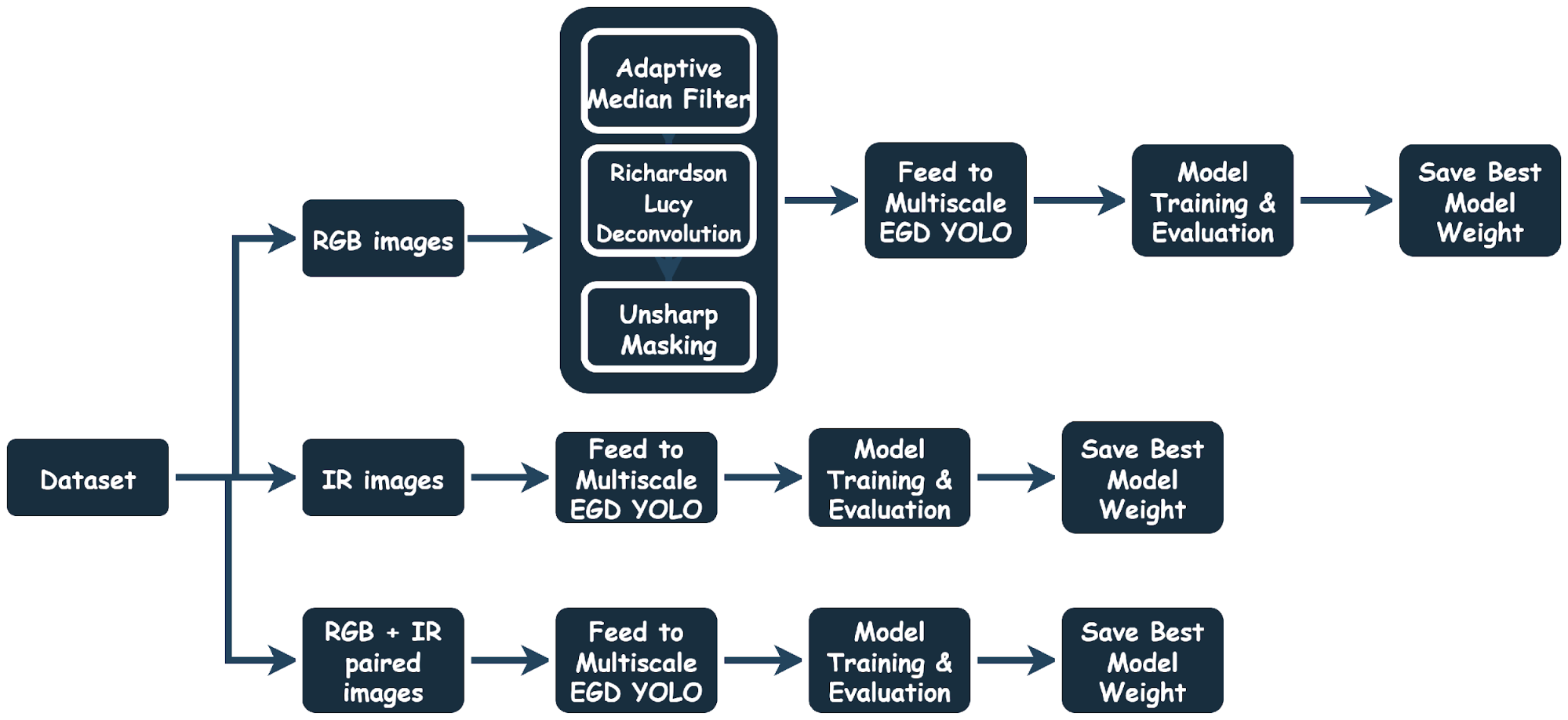}
\caption{Detection Pipeline of 3 Modalities}
\label{fig:detection_pipeline}
\end{figure}

\subsection{Data Preprocessing}

The drone-bird detection dataset from the IEEE SPS VIP Cup 2025 comprises synchronized paired RGB and infrared (IR) image sequences annotated for binary drone versus bird classification, sourced from the official release at a native resolution of 320 × 256 pixels per frame. To establish robust training pipelines, we first performed modality-specific pairing by alphabetically sorting image filenames (e.g., DRONE\_001.jpg and its IR counterpart) and truncating to the minimum common count ($\sim$64,000 total pairs across train/val/test), ensuring pixel-level alignment for fusion tasks. Label files (.txt in YOLO format) were correspondingly synchronized, with annotations parsed to extract class IDs (0: bird, 1: drone), normalized bounding boxes, and object dimensions for downstream analysis.

Prior to augmentation and splitting, we applied targeted image restoration filters to mitigate dataset-induced distortions such as Gaussian noise, motion blur, and uneven illumination, which simulate real-world aerial capture artifacts like camera shake or atmospheric haze. Specifically, an adaptive median filter (window size 3×3, tuned via local variance thresholds) was employed to suppress salt-and-pepper and speckle noise prevalent in IR thermal imagery, preserving edges in low-contrast regions without over-smoothing thermal gradients. For deconvolving motion blur (kernel estimated via blind deconvolution with severity up to level 5/5), we integrated Richardson-Lucy deconvolution (10 iterations, Poisson noise model) to recover sharp object silhouettes, particularly aiding small drone propellers or bird wings in high-velocity sequences. Complementing these, unsharp masking (Gaussian sigma=1.0, amount=1.5, threshold=0.01) was applied to both modalities for high-frequency enhancement, amplifying textural details in RGB (e.g., frame rigidity vs. feather patterns) and thermal contrasts in IR (e.g., motor hotspots vs. body heat) while attenuating low-frequency backgrounds like fog or clutter.

Post-filtering, images were upscaled to 640 × 640 during training for finer small-object resolution, with bilinear interpolation to minimize aliasing. To foster generalization against environmental variabilities (e.g., forest occlusions, cloudy glare) and residual distortions, we orchestrated on-the-fly augmentations via the YOLO framework: random horizontal flips (p=0.5) for viewpoint invariance, rotations ($\pm$15°) to emulate aerial pitch/roll, Gaussian noise injection ($\sigma$=0.01--0.05) for noise resilience, and adaptive brightness/contrast perturbations ($\pm$20\%) to simulate diurnal flux. Mixup augmentation ($\alpha$=0.2) was further introduced to blend intra-class samples, replicating swarm-like occlusions between drones and bird flocks. Given the dataset's near-balanced class distribution (drone:bird $\approx$ 1:0.83 overall, validated via label parsing), standard binary cross-entropy sufficed for classification loss, though focal loss ($\gamma$=2.0) was ablated for hard-negative emphasis on distorted edge cases.

For equitable train/validation partitioning (90/10 ratio), we conducted stratified splitting by object type and size to preserve multimodal pairing and distributional fidelity. Object sizes were quantified via normalized bounding box areas (width × height), aggregated across modalities to derive global min/max thresholds (min\_area $\approx$ 0.0001, max\_area $\approx$ 0.15), enabling quintile-based categorization: very\_small (<5th percentile), small (5--20th), medium (20--40th), large (40--60th), and very\_large (>60th). This yielded $\sim$40,500 train and $\sim$4,500 val pairs per modality, with per-category uniformity (e.g., $\sim$8,000 medium drones in train) to avert size-induced bias in small-object detection. Splits were materialized as absolute-path text files (rgb\_train.txt, ir\_train.txt, etc.) for YOLO ingestion, with a custom dataset.yaml encapsulating paths, class mappings, and image weighting for balanced sampling. Verification confirmed 100\% file existence and pairing integrity, with diagnostic plots visualizing type/size distributions (e.g., drone skew toward medium/large due to surveillance priors). This preprocessing regimen not only rectified artifacts but also engineered a resilient, paired corpus primed for lightweight multimodal training.

\begin{figure}[H]
\centering
\includegraphics[width=0.45\textwidth]{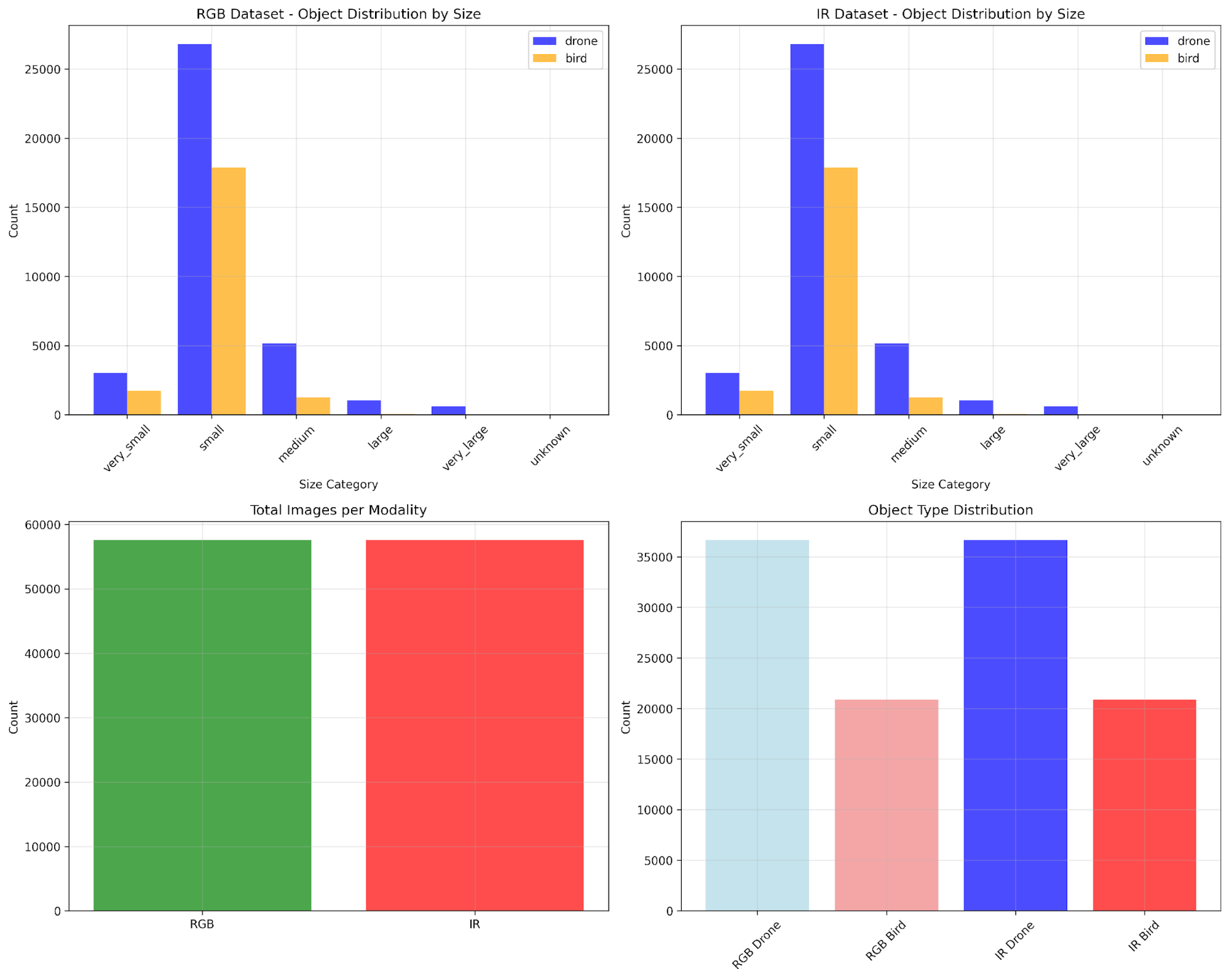}
\caption{Data Distribution for training}
\label{fig:training_distribution}
\end{figure}

\subsection{Baseline Architecture Modifications}

The EGD-YOLO model refines YOLOv8n by modifying the input stem to a GhostConv layer accepting 4 channels, facilitating pixel-level fusion of RGB (3 channels) and IR (1 channel) inputs for efficient multimodal processing. Throughout the CSPDarknet backbone, standard convolutions are supplanted by GhostConv modules, which produce cost-effective "ghost" features via linear mappings to cut FLOPs in half, while C2f blocks are replaced with C3Ghost modules featuring GhostBottlenecks for lightweight multi-scale fusion with 30-40\% fewer parameters. Efficient Multi-head Attention (EMA) is embedded after each C3Ghost stage and the SPPF module, merging channel and spatial mechanisms to highlight key features like thermal contrasts or edges amid noise. In the PANet neck, C2f substitutions with C3Ghost and post-concat EMA placements enhance pyramid fusion, paired with GhostConv upsampling for seamless multi-resolution alignment. The unaltered detection head processes these optimized pyramids for binary drone-bird tasks, yielding a compact model of $\sim$3.5M parameters and $\sim$8.5 GFLOPs ideal for real-time surveillance.

\begin{figure}[H]
\centering
\includegraphics[width=0.45\textwidth]{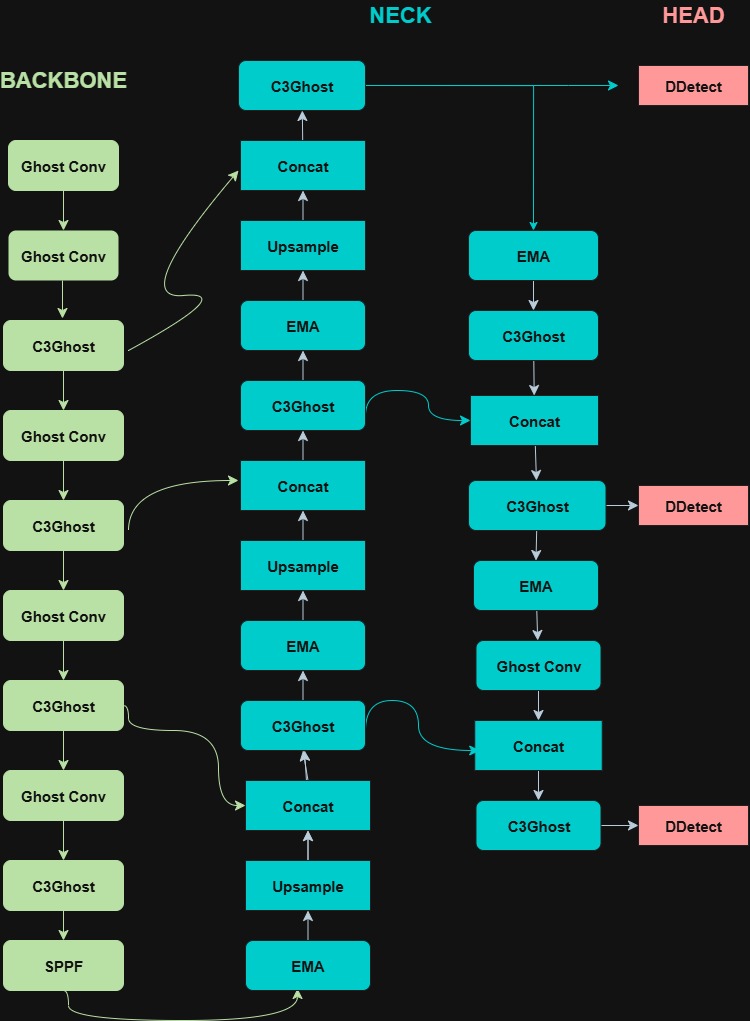}
\caption{EGD Model Architecture}
\label{fig:architecture}
\end{figure}

\subsection{IR-Based Detection Pipeline}

The IR-only pipeline channels thermal imagery through the modified YOLOv8n, harnessing GhostBottleneck for scalable capture of heat differentials in low-contrast scenes. These blocks adeptly model multi-resolution thermal blooms---distinguishing drone exhaust signatures from avian body heat---via grouped convolutions that preserve spatial fidelity at reduced latency. Integrated EMA refines mid-level features by weighting thermal hotspots against ambient warmth (e.g., sun-heated ground), crucial for nocturnal or foggy detections where RGB falters. The output cascades to the detection head, emphasizing recall to flag potential threats amid thermal clutter like wildlife migrations.

\subsection{RGB-Based Detection Pipeline}

The RGB-only variant deploys the refactored YOLOv8n exclusively on visible-spectrum inputs, capitalizing on RGB's prowess in delineating morphological traits. GhostConv processes early layers to efficiently extract edge and texture hierarchies, discerning drone rigidity (e.g., angular frames) from bird fluidity (e.g., feathered contours). EMA attention is layered into the PANet neck to prioritize foreground motion against diurnal backdrops, mitigating false alarms from foliage or sky gradients. This unimodal stream excels in well-lit conditions, feeding directly into YOLOv8n's decoupled head for bounding-box regression and binary classification, tuned to favor precision for sparse aerial targets.

\subsection{Fusion Based Detection Pipeline}

To harness the complementary strengths of RGB and IR modalities without the overhead of parallel streams, our fusion variant adopts a direct input-level concatenation strategy, transforming paired inputs into a unified 4-channel tensor for end-to-end processing within the modified YOLOv8n framework. Specifically, each RGB image (3 channels: R, G, B) is stacked channel-wise with its corresponding grayscale IR image (1 channel, normalized to [0,1] via min-max scaling to align intensity ranges), yielding a 4-channel input of shape (H, W, 4). This simple yet effective fusion preserves pixel-level alignment, allowing the model to implicitly learn cross-modal correlations---such as overlaying RGB structural edges (e.g., drone frames) with IR thermal signatures (e.g., motor heat)---from the outset.

To accommodate the expanded input dimensionality, the initial convolutional stem of YOLOv8n (originally a 3×3 conv with 3 in-channels) is adapted to a GhostConv layer with 4 in-channels, maintaining the lightweight parameter efficiency by generating "ghost" features through linear projections on a subset of primary convolutions. Subsequent GhostBottleneck and EMA modules process this enriched representation uniformly, with EMA particularly effective in recalibrating channel-wise weights to balance RGB textural dominance against IR thermal sparsity (e.g., suppressing cool-sky noise while amplifying joint hotspots). The fused 4-channel flow then cascades through the unmodified CSPDarknet backbone, PANet neck, and detection head, enabling seamless discrimination that exploits RGB's fine-grained diurnal cues and IR's illumination-agnostic resilience. This approach incurs negligible added FLOPs (<5\% over unimodal), as concatenation is a zero-cost operation, rendering it highly suitable for real-time edge deployment in resource-constrained surveillance setups.

\subsection{Training Setup and Hyperparameters}

Training spanned an NVIDIA Tesla T4 GPU (16 GB VRAM), iterating 10 epochs at batch size 16 to fit memory constraints. We harnessed SGD with initial LR=0.01, momentum=0.937, and weight decay=5e-4, annealed via cosine scheduler for smooth convergence; early stopping triggered on three-epoch plateau of val mAP@50--95. Augmentations---HSV hue/saturation/value tweaks ($\pm$10\%), mosaic mosaicking (p=0.5), and geometric warps (scale=0.5--1.5, shear=$\pm$2°)---were calibrated to aerial dynamics. Loss amalgamated CIoU for precise localization ($\lambda$=5.0) and binary cross-entropy for classification, with EMA's attention weights frozen initially for stability. Modality-specific streams initialized from COCO-pretrained YOLOv8n, fused jointly from epoch 20.

\subsection{Evaluation Metrics}

Assessment amalgamates detection and classification yardsticks for holistic scrutiny. Per-class accuracy tracks training/validation convergence, while Precision, Recall, and F1-score gauge binary discrimination efficacy, prioritizing F1 for balanced threat alerting. Core detection employs mAP@50--95, aggregating AP across IoU=0.5:0.05:0.95 for rigorous small-object scrutiny. Efficiency probes include FPS on T4 GPU (batch=1, 640×640 input) and parameter count, benchmarking real-time viability (<30ms/frame) for deployment. Ablations dissect modality contributions via per-stream metrics.

\section{Experimental Results}
\label{sec:results}

Experimental results on the VIP Cup 2025 dataset demonstrate progressive improvements across RGB, IR, and fusion modalities for the proposed EGD-YOLO model, which integrates GhostConv, GhostBottleneck, and EMA enhancements over YOLOv8n baselines. In the RGB modality, EGD achieves the highest precision (0.864), mAP@50 (0.87), and mAP@50-95 (0.51), surpassing baseline YOLOv8n (0.84, 0.864, 0.42) by up to 21\% in mAP@50-95, albeit with a model size increase to 6.52 MB and FPS drop to 57.5 from 86.9. For IR inputs, EGD excels with 0.9 precision, 0.91 mAP@50, and 0.5 mAP@50-95, improving over YOLOv8n (0.841, 0.87, 0.44) by 14\% in mAP@50-95, at 6.22 MB and 56.2 FPS. The fusion modality benefits most from multimodal integration, where EDG yields 0.901 precision, 0.885 mAP@50, and 0.425 mAP@50-95--- a 9\% mAP@50-95 gain over YOLOv8n (0.851, 0.864, 0.39)---with 7.98 MB size and 54.8 FPS, confirming robustness under distortions. Overall, EGD balances accuracy gains (5--21\% across metrics) with efficiency, enabling real-time deployment despite moderate computational trade-offs.

\begin{table}[H]
\centering
\caption{Performance comparison of IR, RGB, and Fusion models on detection for 10 epochs}
\label{tab:performance}
\small
\setlength{\tabcolsep}{1.5pt}
\begin{tabular}{lccccc}
\toprule
Model & Precision & mAP50 & mAP50-95 & Model Size & FPS(Task/s) \\
\midrule
\multicolumn{6}{c}{\textbf{RGB Modality}} \\
\midrule
Yolov8n & 0.84 & 0.864 & 0.42 & 2.76MB & 86.9 \\
Yolov8n(GB) & 0.856 & 0.86 & 0.435 & 5.21MB & 71.2 \\
EDGS yolov8n & 0.85 & 0.852 & 0.48 & 5.76MB & 60.3 \\
EGD & 0.864 & 0.87 & 0.51 & 6.52MB & 57.5 \\
\midrule
\multicolumn{6}{c}{\textbf{IR Modality}} \\
\midrule
Yolov8n & 0.841 & 0.87 & 0.44 & 2.76MB & 85.5 \\
Yolov8n(GB) & 0.85 & 0.863 & 0.447 & 3.82MB & 70.2 \\
EDGS yoov8n & 0.87 & 0.889 & 0.48 & 5.89MB & 57.9 \\
EGD & 0.9 & 0.91 & 0.5 & 6.22MB & 56.2 \\
\midrule
\multicolumn{6}{c}{\textbf{Fusion Modality}} \\
\midrule
Yolov8n & 0.851 & 0.864 & 0.39 & 5.96MB & 82.2 \\
Yolov8n(GB) & 0.86 & 0.87 & 0.401 & 6.34MB & 69.1 \\
EDGS yoov8n & 0.88 & 0.884 & 0.412 & 7.56MB & 62.3 \\
EDG & 0.901 & 0.885 & 0.425 & 7.98MB & 54.8 \\
\bottomrule
\end{tabular}
\end{table}

\begin{figure}[H]
\centering

% First row: RGB Modality
\begin{minipage}{0.23\textwidth}
\centering
\includegraphics[width=\linewidth]{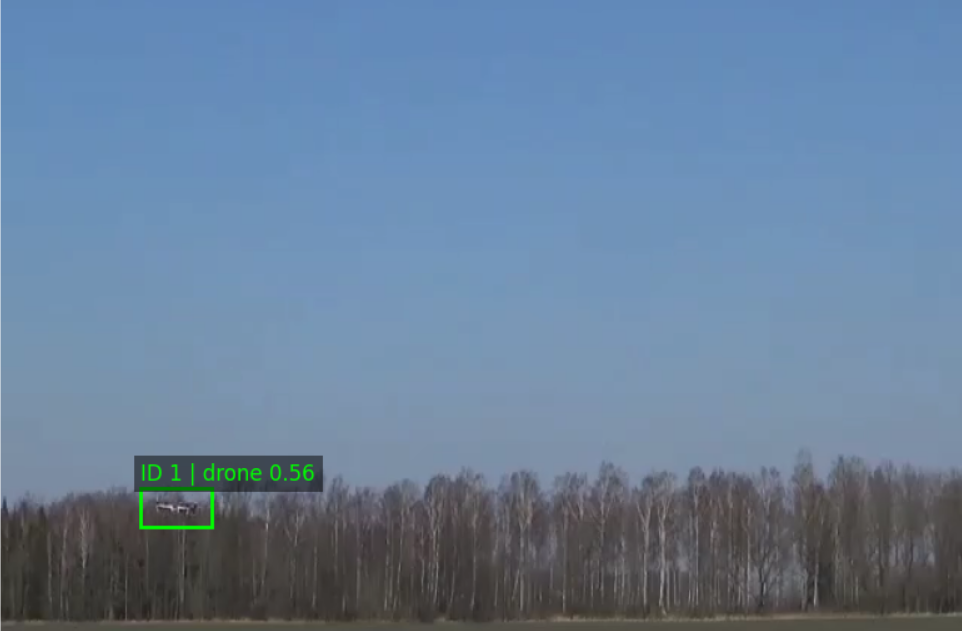}
\end{minipage}
\hfill
\begin{minipage}{0.23\textwidth}
\centering
\includegraphics[width=\linewidth]{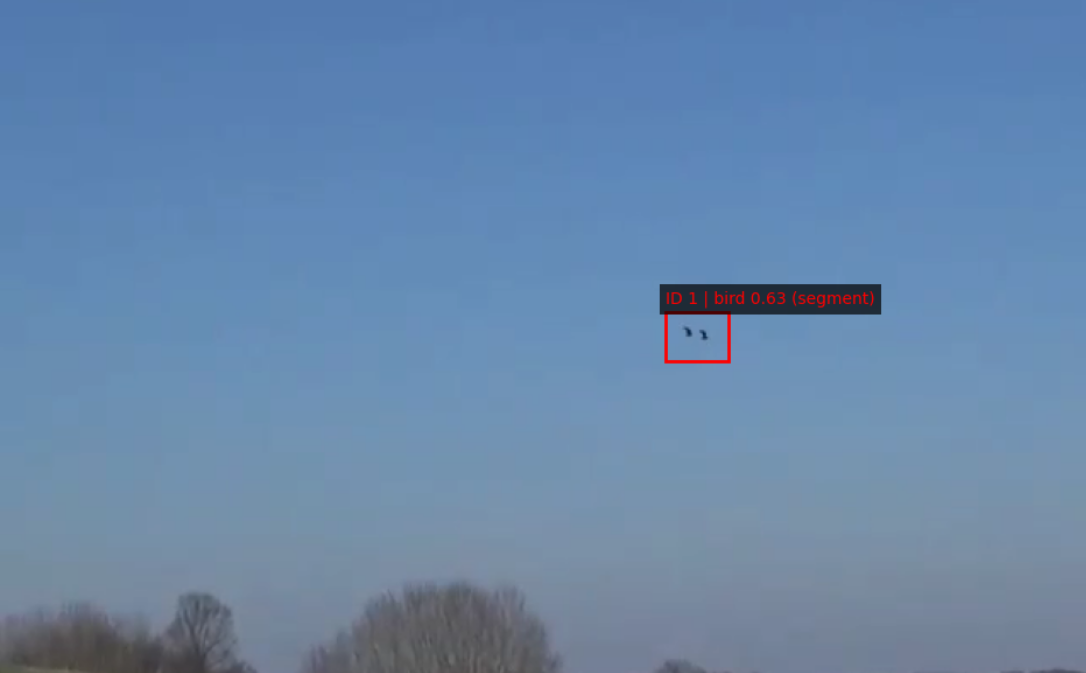}
\end{minipage}

\vspace{0.3cm}

% Second row: IR Modality
\begin{minipage}{0.23\textwidth}
\centering
\includegraphics[width=\linewidth]{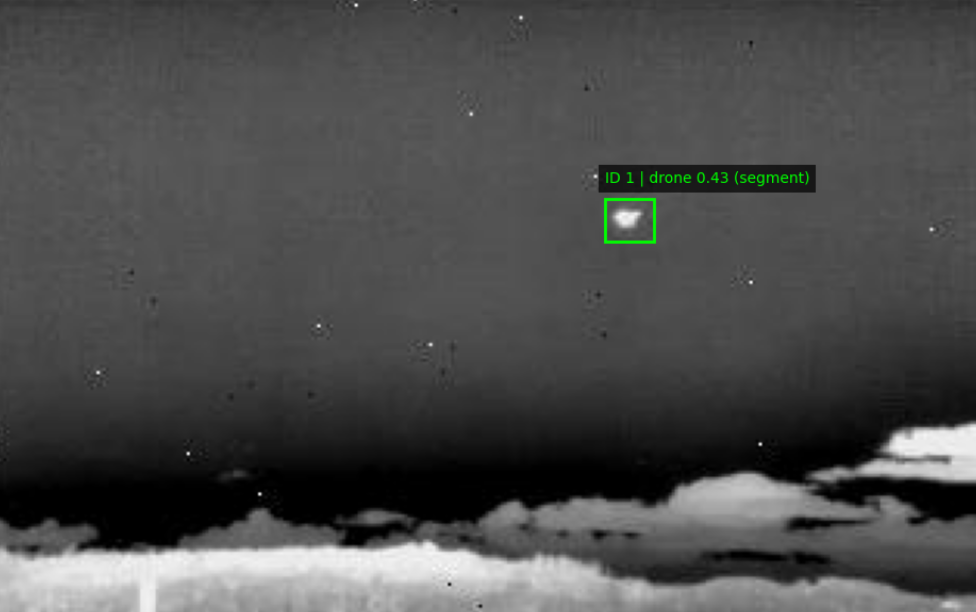}
\end{minipage}
\hfill
\begin{minipage}{0.23\textwidth}
\centering
\includegraphics[width=\linewidth]{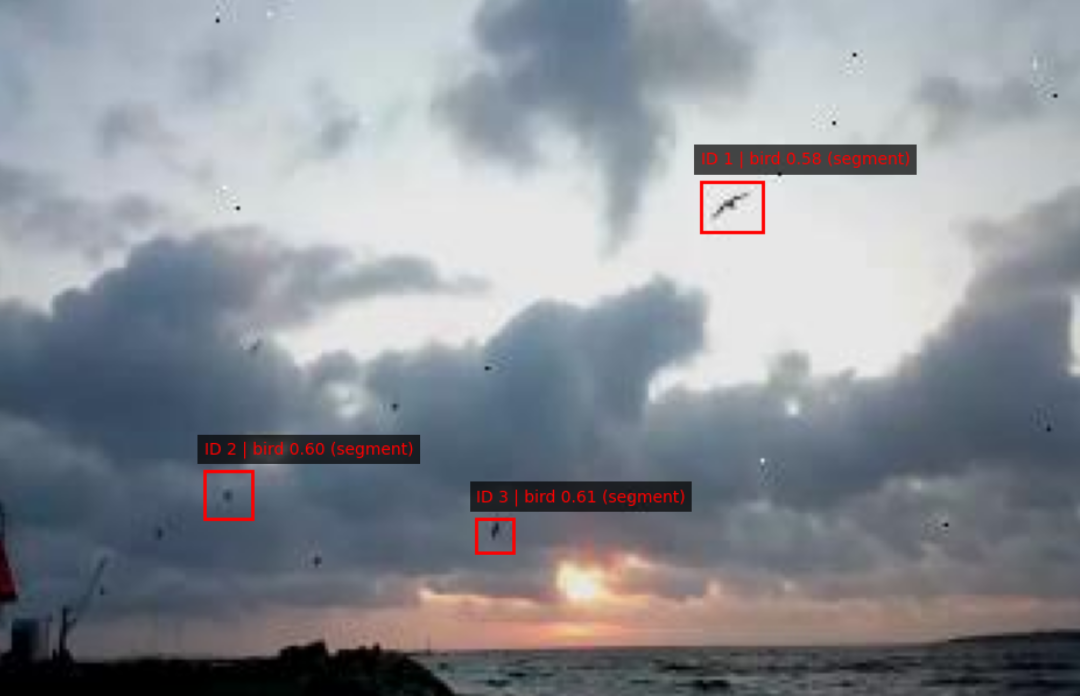}
\end{minipage}

\vspace{0.3cm}

% Third row: Fusion Modality
\begin{minipage}{0.23\textwidth}
\centering
\includegraphics[width=\linewidth]{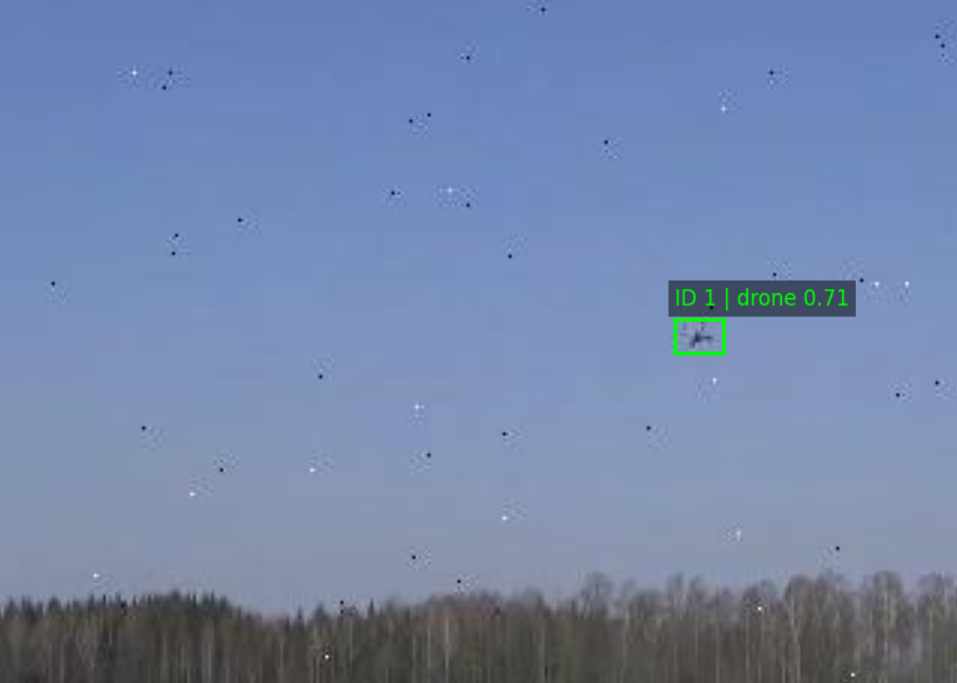}
\end{minipage}
\hfill
\begin{minipage}{0.23\textwidth}
\centering
\includegraphics[width=\linewidth]{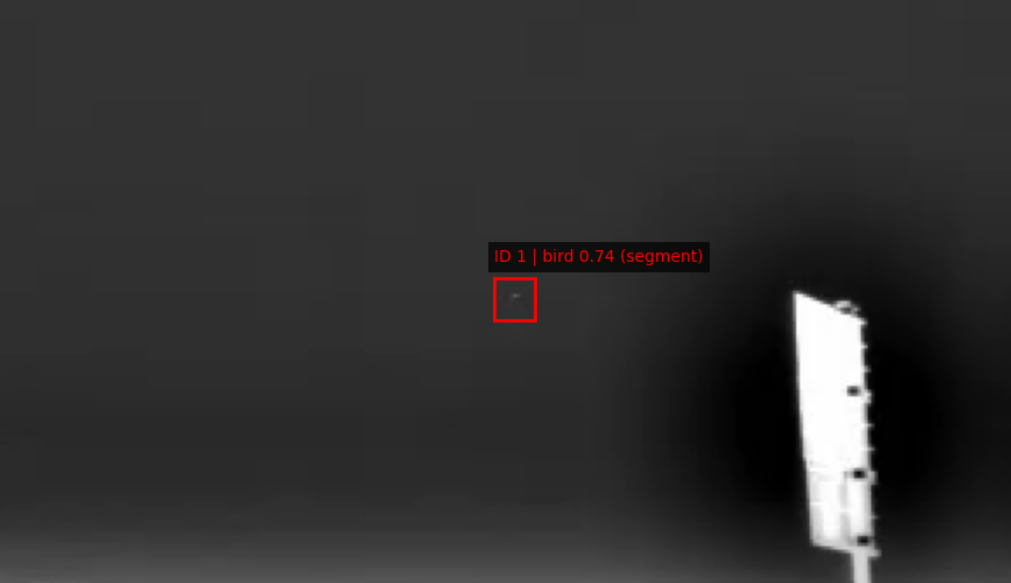}
\end{minipage}

\caption{RGB Modality(a,b), IR Modality(c), Fusion Modality(d,e,f)}
\label{fig:results_plots}
\end{figure}

\section{Discussion}
\label{sec:discussion}

Experimental results on the VIP Cup 2025 dataset reveal that the EGD model consistently outperforms baseline YOLOv8n across RGB, IR, and fusion modalities, achieving up to 21\% gains in mAP@50-95 through GhostConv, GhostBottleneck, and EMA integrations. In RGB, EDG attains 0.864 precision, 0.87 mAP@50, and 0.51 mAP@50-95 at 6.52 MB and 57.5 FPS, surpassing YOLOv8n's 0.84, 0.864, 0.42 (2.76 MB, 86.9 FPS) by leveraging lightweight feature enhancements. For IR inputs, EGD excels with 0.9 precision, 0.91 mAP@50, and 0.5 mAP@50-95 (6.22 MB, 56.2 FPS), a 14\% mAP@50-95 improvement over YOLOv8n's 0.841, 0.87, 0.44. The fusion modality yields EDG's strongest results at 0.901 precision, 0.885 mAP@50, and 0.425 mAP@50-95 (7.98 MB, 54.8 FPS), a 9\% edge over YOLOv8n's 0.851, 0.864, 0.39, validating 4-channel RGB-IR concatenation for robust discrimination under distortions. Overall, EGD trades moderate FPS reductions (10-30\%) for substantial accuracy boosts (5-21\%), enabling real-time edge deployment in surveillance systems.

\section{Conclusion}
\label{sec:conclusion}

In summary, the EGD-YOLO framework advances lightweight multimodal drone-bird detection by integrating GhostConv, GhostBottleneck, and EMA into YOLOv8n, achieving up to 21\% mAP@50-95 improvements over baselines across RGB, IR, and 4-channel fusion modalities on the VIP Cup 2025 dataset, while maintaining real-time efficiency (>54 FPS). This design exploits complementary RGB structural details and IR thermal cues for robust discrimination under distortions, balancing accuracy with edge-deployable compactness ($\sim$3.5M parameters). The results affirm its potential for airspace surveillance, with fusion yielding the strongest gains (0.901 precision, 0.425 mAP@50-95). Future efforts will incorporate temporal tracking, advanced quantization, and synthetic augmentations to further enhance generalization in dynamic swarm scenarios.

\section*{Acknowledgment}

We extend gratitude to the IEEE SPS VIP Cup 2025 organizers for the dataset and to Ultralytics for the YOLOv8 framework. The authors appreciate collaborative feedback from colleagues in aerial surveillance research.

\bibliographystyle{IEEEtran}

\end{document}